\documentclass[10pt,twocolumn,letterpaper]{article}
 
\usepackage{iccv}
\usepackage{times}
\usepackage{epsfig}
\usepackage{graphicx}
\usepackage{amsmath}
\usepackage{amssymb}
\usepackage{multirow}
\usepackage{booktabs}
\usepackage{array}
\usepackage{paralist}
\usepackage{tabularx}
\usepackage{cuted}
\usepackage{capt-of}
\usepackage{hyperref}
\iccvfinalcopy

\begin{document}
\title{Diverse Image Inpainting with Bidirectional and Autoregressive Transformers}
\author{Yingchen Yu$^{1}$\thanks{Equal contribution}\quad \quad
Fangneng Zhan$^1$\footnotemark[1]\quad \quad
Rongliang Wu$^1$\quad \quad
Jianxiong Pan$^2$\quad \quad
Kaiwen Cui$^1$ \\
Shijian Lu$^1$\thanks{Corresponding author}\quad \quad
Feiying Ma$^2$\quad \quad
Xuansong Xie$^2$\quad \quad
Chunyan Miao$^1$\\
$^1$ Nanyang Technological University $\ \
^2$ DAMO Academy, Alibaba Group\\
{\tt\small \{yingchen001, ronglian001, kaiwen001\}@e.ntu.edu.sg, \{fnzhan, shijian.lu, ascymiao\}@ntu.edu.sg} \\
{\tt\small \{jianxiong.pjx, feiying.mfy\}@alibaba-inc.com, xingtong.xxs@taobao.com} 
}
\maketitle
\pagestyle{empty}  
\thispagestyle{empty} 
\begin{strip}\centering
\vspace{-5.5em}
\includegraphics[width=\textwidth]{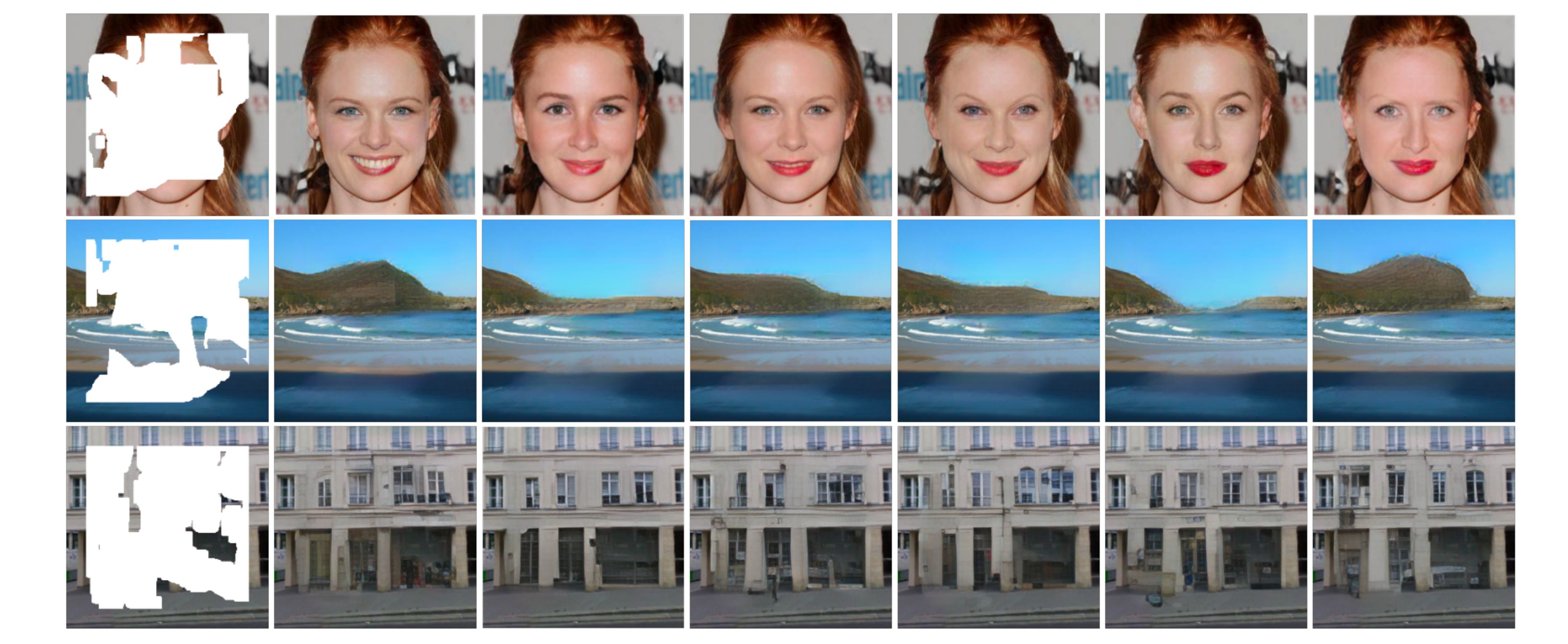}
\captionof{figure}{
The proposed BAT-Fill introduces a novel bidirectional autoregressive transformer that captures deep bidirectional contexts for autoregressive generation of diverse contents in image inpainting. Evaluations over multiple public datasets show that BAT-Fill can generate realistic and reasonable image contents. The three illustrative sample images from top to bottom are selected from the datasets CelebA-HQ \cite{karras2017progressive}, Places2 \cite{zhou2017places}, and Paris StreetView \cite{pathak2016context}, respectively. 
\label{fig_teaser}}
\end{strip}

\begin{abstract}
Image inpainting is an underdetermined inverse problem, which naturally allows diverse contents to fill up the missing or corrupted regions realistically. Prevalent approaches using convolutional neural networks (CNNs) can synthesize visually pleasant contents, but CNNs suffer from limited perception fields for capturing global features. With image-level attention, transformers enable to model long-range dependencies and generate diverse contents with autoregressive modeling of pixel-sequence distributions. However, the unidirectional attention in autoregressive transformers is suboptimal as corrupted image regions may have arbitrary shapes with contexts from any direction. We propose BAT-Fill, an innovative image inpainting framework that introduces a novel bidirectional autoregressive transformer (BAT) for image inpainting. 
BAT utilizes the transformers to learn autoregressive distributions, which naturally allows the diverse generation of missing contents. In addition, it incorporates the masked language model like BERT, which enables bidirectionally modeling of contextual information of missing regions for better image completion. Extensive experiments over multiple datasets show that BAT-Fill achieves superior diversity and fidelity in image inpainting qualitatively and quantitatively.
\end{abstract}

\maketitle

\section{Introduction}
As an ill-posed problem, image inpainting naturally allows numerous solutions as long as the restored images are realistic and semantically reasonable as illustrated in Fig.~\ref{fig_teaser}. However, it remains a great challenge to synthesize diverse while realistic contents that maintain integrity and consistency with the uncorrupted image regions, especially when the corrupted image regions are large and rich in complex textures and structures.

Recently, GAN-based (generative adversarial network) inpainting ~\cite{pathak2016context, yu2019free, nazeri2019edgeconnect, Liu2019MEDFE} has achieved remarkable progress by training with reconstruction and adversarial losses over large-scale datasets. However, these methods are trained to learn the one-to-one mapping from masked images to complete images, which results in the incapacity of producing diverse inpainting results. 
In contrast to deterministic inpainting, a few studies~\cite{zheng2019pluralistic,zhao2020uctgan} attempt for diverse inpainting with variational auto-encoder (VAE) networks \cite{kingma2013auto}, but the inpainting quality is often compromised while generating complex structural and texture patterns due to the limited capacity of parametric distributions~\cite{zhao2017towards}. Instead of parametric distribution modeling like VAE-based methods, \cite{peng2021generating} utilizes a CNN-based conditional network to learn an autoregressive distribution for recovering diverse and structural features. As the autoregressive models are optimized to encode unidirectional context only, which means the informative contexts of valid pixels after the current position are substantially ignored.
To explore bidirectional context, \cite{wan2021high} adopts the masked language model (MLM) like BERT \cite{devlin2018bert}. However, MLM predicts the masked tokens independently which may oversimplify the complex context dependency in the data \cite{song2020mpnet} and result in inconsistency in the generated results. 

In this paper, we propose a bidirectional and autoregressive transformer (BAT) that marries the best of autoregressive modeling and MLM to model deep bidirectional contexts in an autoregressive manner. In the proposed BAT, we permute the input sequence by sorting the valid and missing pixels and start autoregressive modeling at the position of the first missing pixel. With all available contexts in front, BAT can exploit bidirectional contexts and spatial dependency simultaneously. 
In addition, we adopt the two-stage completion procedure as reported in \cite{wan2021high} and develop BAT-Fill, an image inpainting network that firstly recovers the diverse yet coherent image structures based on the proposed BAT and then exploits a CNN-based texture generator to up-sample the coarse structures and synthesize texture details. Extensive experiments show that BAT-Fill achieves superior image inpainting performance. 

The main contributions of this work can be summarized in three aspects. 
First, we adopt the transformers 
to learns an autoregressive distribution for diverse image inpainting, which effectively improves the modeling capacity for long-range dependencies and global structures.
Second, we design a novel bidirectional and autoregressive transformer (BAT) that captures bidirectional information and establishes output dependency simultaneously.
Third, extensive experiments over multiple datasets show that the proposed method achieves superior performance as compared with the state-of-the-art in both inpainting quality and inpainting diversity.

\section{Related Work}

\subsection{Image Inpainting}

As an ill-posed problem, realistic and high-fidelity image inpainting is a challenging task that has been studied for years.
Based on the inpainting outcome, most existing image inpainting 
methods can be broadly classified into two categories including deterministic image inpainting and diverse image inpainting.

\subsubsection{Deterministic Image Inpainting}
Traditional methods address image inpainting challenge through either image diffusion \cite{bertalmio2000image, ballester2001filling} or using image patches \cite{barnes2009patchmatch, hays2007scene, darabi2012image}. However, diffusion-based methods often introduce diffusion-related blurs, which tends to fail while the missing or corrupted image regions are large \cite{bertalmio2003simultaneous,ballester2001variational, bertalmio2001navier}. Patch-based methods can work well for the inpainting of stationary background with repeating patterns. However, they struggle in completing large missing regions of complex scenes as the patch-based approach relies heavily on patch-wise matching of low-level features.

Generative adversarial networks (GANs) \cite{goodfellow2014generative}
have been investigated extensively in various image synthesise tasks such as image translation \cite{park2019semantic,shrivastava2017simgan,isola2017image,zhan2019gadan,zhan2019esir,zhan2020sagan,zhan2021unite},
image editing \cite{yu2018inpainting,wu2020cascade,wu2020leed}, image composition \cite{lin2018st,zhang2021defect,zhan2019sfgan,zhan2020aicnet,zhan2020emlight,zhan2021gmlight}, etc. 
Specifically for image inpainting, Pathak \etal \cite{pathak2016context} first apply adversarial learning to the image inpainting task. To further improve the adversarial learning within local regions, Iizuka \etal \cite{iizuka2017globally} introduce an extra local discriminator to enforce the local consistency. As the local discriminator uses fully-connected layers and can only deal with missing regions of fixed shapes, Yu \etal \cite{yu2018generative} inherit the discriminator from PatchGAN \cite{isola2017image} due to its great success in image translation. 
Yan \etal \cite{yan2018shift} propose patch-swap to make use of distant feature patches for the better inpainting quality. 
Liu \etal \cite{liu2018image} design partial convolutions to alleviate the negative influence of the masked regions. Yu \etal \cite{yu2019free} present a novel free-form image inpainting system based on an end-to-end generative network with gated convolutions. To generate reasonable structures and realistic textures, Nazeri \etal \cite{nazeri2019edgeconnect} and Xu \etal \cite{xu2020e2i} utilize edge maps as structural guidance for image inpainting, and Ren \etal \cite{ren2019structureflow} instead propose to use edge-preserved smooth images as structural guidance. Liu \etal \cite{liu2020rethink} propose feature equalizations to improve the consistency between structures and textures. As aforementioned methods focus on reconstructing the ground truth instead of generating pluralistic inpainting, they are constraint to generate a deterministic inpainting image for each incomplete image.

\begin{figure*}[t]
\centering
\includegraphics[width=1.0\linewidth]{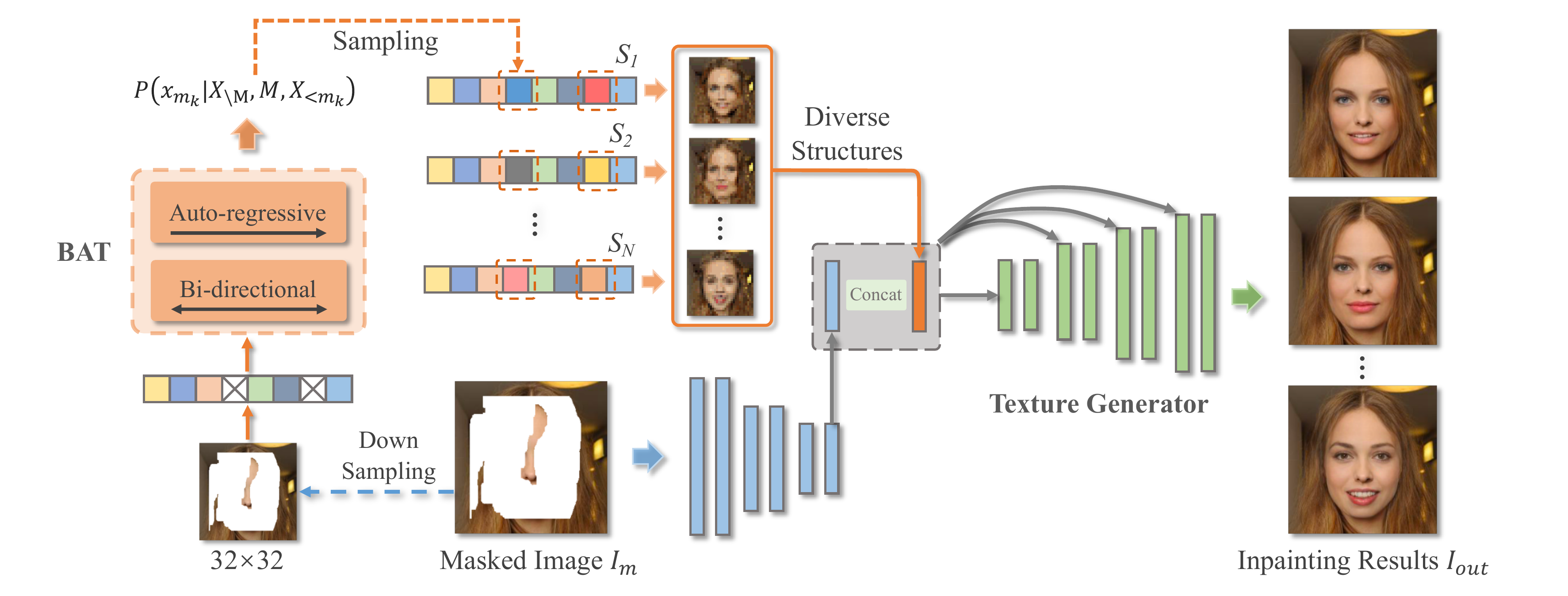}
\caption{
Overview of the proposed image inpainting method: Given a $Masked$ $Image$ $I_m$, the proposed $BAT$-$Fill$ first $Down$ $Sample$ it to a lower resolution and feed the down-sampled image to the $Bidirectional$  $Autoregressive$ $Transformer (BAT)$ for the recovery of $Diverse$ $Structures$. Taking the recovered image structures and image style features (extracted by the $Encoder$) as inputs, the $Texture$ $Generator$ synthesizes high-resolution texture and produces the $Inpainting\ Results$ $I_{out}$.
}
\label{fig_arch}
\end{figure*}

\subsubsection{Diverse Image Inpainting}
To achieve pluralistic image inpainting with plausible filling contents, Zheng \etal \cite{zheng2019pluralistic} propose a VAE-based network with a dual pipeline, which trades off between reconstructing ground truth and maintaining the diversity of the inpainting results.
Similarly, Zhao \etal \cite{zhao2020uctgan} propose a VAE-based model and leverage a reference image to improve the diversity. Although the above methods achieve certain diversities to some extent, completion quality of VAE-based methods is limited due to variational training. 
Recently, Zhao \etal \cite{zhao2021comodgan} propose a co-modulated GAN to incorporate the image condition and the stochastic generation of unconditional generative model for diverse inpainting.
Peng \etal \cite{peng2021generating} introduce a hierarchical vector quantized variational auto-encoder (VQ-VAE) to quantize the context representation and archive diverse structure generation in an autoregressive way. 
Sharing a similar framework with us, Wan \etal \cite{wan2021high} propose to apply transformer for diverse structure generation using the objective of BERT~\cite{devlin2018bert}. In contrast, we propose a novel Bidirectional and Autoregressive Transformer (BAT) which inherits the advantages of autoregressive models and bidirectional models and achieves superior image inpainting performance.

\subsection{Transformers in Vision}

Transformer has emerged as a powerful tool to model the interactions between sequences regardless of the relative position.
Specially, 
Vaswani \etal \cite{vaswani2017attention} employ transformers for image classification by treating an image as a sequence of patches.  
DETR \cite{carion2020end} utilizes transformer decoder to model object detection as an end-to-end dictionary lookup problem with learnable queries,
thus removing the hand-crafted processes such as Non-Maximal Suppression (NMS). 
Based on DETR, deformable DETR \cite{zhu2020deformable} further introduces a deformable attention layer to focus on a sparse set of contextual elements which achieves fast convergence and better detection performance.
Recently, Vision Transformer (ViT) \cite{dosovitskiy2020image} showed that pure-transformer networks can also achieve excellent image classification performance as compared with CNN-based methods.
DeiT \cite{touvron2020training} further extends ViT by introducing a novel distillation approach. 
BoTNet \cite{srinivas2021bottleneck} replaces the spatial 3$\times$3 convolution layers with multi-head self-attention in certain stages of the original ResNet \cite{he2016deep}, demonstrating very competitive performance on different visual recognition tasks.
Esser \etal \cite{esser2020taming} adapt transformers and VQ-VAE in both conditional and unconditional generation tasks, and achieve high-fidelity synthesis of megapixel images. 

Instead of leveraging features of transformers for high-level tasks or generate pixels autoregressively, we specifically propose a novel Bidirectional and Autoregressive Transformer (BAT) for image inpainting, so that the model can learn both bidirectional context and output dependency. 

\section{Proposed Method}

As illustrated in Fig.~\ref{fig_arch}, the proposed BAT-Fill consists of two major parts including a diverse-structure generator for the reconstruction of coarse image structures and a texture generator for the generation of fine-grained texture details. The diverse-structure generator incorporates and adapts a transformer architecture that models the distribution of global structural information and recovers complete and coherent low-resolution structures $S_1, S_2, \cdots, S_N$ given a \textit{Masked Image} $I_m$ as input. 
Under the guidance of coarse structure $S_i, i \in [1, N]$ and corrupted image $I_m$, the \textit{Texture Generator} synthesizes high-resolution fine-grained texture to produce the \textit{Inpainting Results} $I_{out}$.
Once the full model is trained, we can sample different image structures $S_i, i \in [1, N]$ by the diverse-structure generator and thus generate diverse inpainting results with the texture generator, more details to be discussed in the ensuing subsections.

\subsection{Diverse-structure Generator}

\subsubsection{Context Representation}
To relieve the pressure of quadratic complexity incurred in transformer, we adapt the low-resolution image with the size of $32\times32\times3$ to represent the coarse structure. As the autoregressive generation requires discrete distribution, the pixel value should be treated as classes to the model, which leads to the dimensionatliy of $256^3$ for each pixel of the 8-bit RGB images. Following Chen \etal \cite{chen2020generative}, a color palette is applied to further reduce the dimensionality to $512$ while faithfully preserving the main structure of original images, which is generated by $k$-means clustering of RGB pixel values with $k$=512 from ImageNet~\cite{deng2009imagenet} dataset.

\begin{figure*}[t]
\centering
\includegraphics[width=1.0\linewidth]{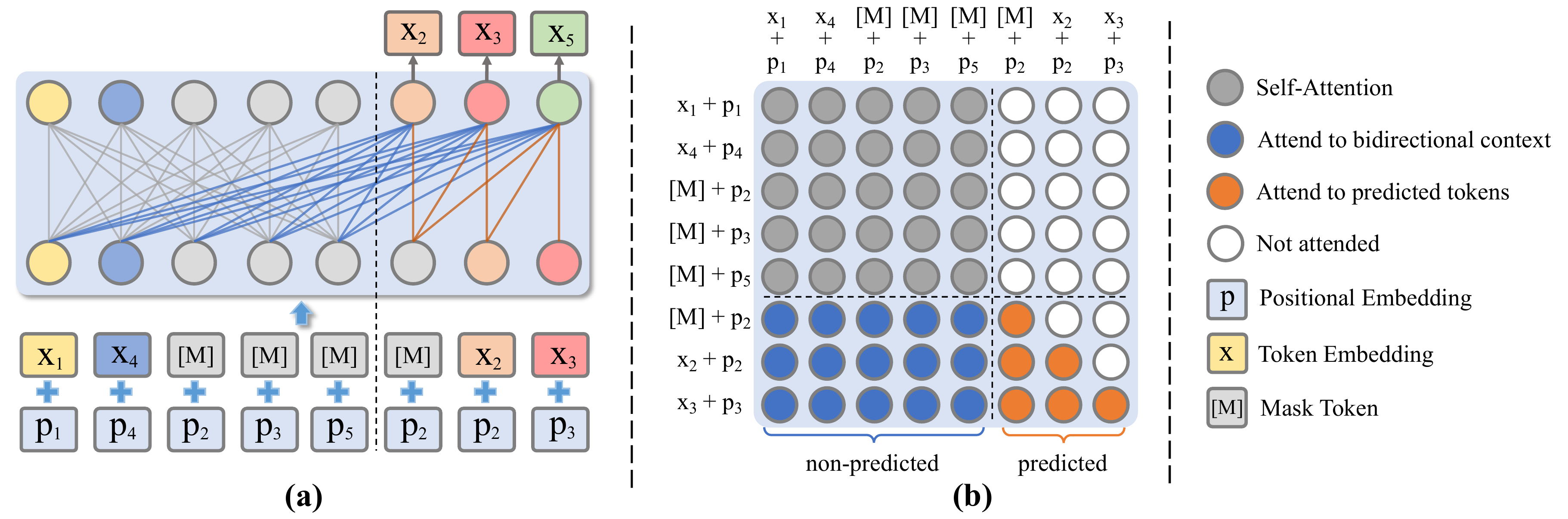}
\caption{
Illustrations of $BAT$ and its attention mask. 
(a) All unmasked tokens (\ie $X_1, X_4$) is permuted to the front of the sequence which is followed by masked tokens with positional embedding (\ie $p_2, p3, p5$). It provides bidirectional context for the autoregressive modeling of masked tokens (\ie $X_2, X_3, X_5$).
(b) $BAT$ allows to attend non-predicted tokens and previously predicted tokens for the prediction of masked future tokens. For example, while predicting $X_3$, the model could attend to $X_1, X_4$ (non-predicted tokens) and $X_2$ (previously predicted token) simultaneously. Meanwhile, the future tokens are not attended to prevent information leakage in autoregressive modeling.
}
\label{fig_att}
\end{figure*}

\subsubsection{Bidirectional and Autoregressive Transformer}
Autoregressive (AR) modeling and masked language modeling (MLM) in BERT~\cite{devlin2018bert} are two representative objectives for exploiting large language corpora in language processing tasks. Given a discrete sequence $X = \left \{ x_1, x_2,\dots,x_L\right \}$ where $L$ is the length of $X$, AR model is optimized by maximizing the unidirectional likelihood:
\begin{equation}
\log P(x;\theta) = \mathbb{E}_{X} \sum\limits_{t=1}^{L} \log P(x_{t}|X_{<t};\theta),
\label{eq_AR}
\end{equation}
where $\theta$ is the parameters of the model. In contrast, MLM aims to reconstruct corrupted data with the masked positions $M = \left \{m_1,m_2,\dots,m_K\right\}$, where $K$ is the number of masked tokens. Each masked position of the corrupted data is indicated by a special token $[M]$ following BERT~\cite{devlin2018bert}. Denoting the masked tokens as $X_M$ and unmasked tokens as $X_{\setminus M}$, the objective of MLM can be formulated by:
\begin{equation}
\log P(X_M|X_{\setminus M};\theta) = \mathbb{E}_{X} \sum\limits_{m_k \in M} \log P(x_{m_k}|X_{\setminus M};\theta).
\label{eq_MLM}
\end{equation}
AR and MLM differ from two aspects as defined in Eqs. \ref{eq_AR} and \ref{eq_MLM}. The first aspect lies with \textit{output dependency}, where MLM predicts the masked tokens separately and independently which may oversimplify the complex context dependency in the data~\cite{song2020mpnet}.
As a comparison, AR factorizes the predicted tokens with the product rule, which establishes the output dependency and produces better predictions. The second aspect lies with \textit{context dependency}, where AR is only conditioned on the tokens up to the current position (in a fixed order), while MLM has access to bidirectional contextual information. Therefore, MLM is more suitable for image inpainting as the missing or corrupted image regions often have arbitrary shapes with rich variation in the neighboring background.

We propose a novel Bidirectional and Autoregressive Transformer (BAT) that inherits the advantages of AR and MLM to achieve bidirectional context modeling and output dependency simultaneously. The training objective of the BAT is formulated by:
\begin{equation}
\mathcal{L}_{BAT} = \mathbb{E}_{X} \sum\limits_{m_k \in M} \log P(x_{m_k}|X_{\setminus M},M,X_{<m_k};\theta).
\end{equation}
We first project all the tokens into a $d$-dimensional token embedding and add a learnable position embedding over the token embedding to preserve the positional information. Unlike XLNet~\cite{yang2019xlnet} which randomly permutes the input sequence to capture the bidirectional context, we permute all unmasked tokens $X_{\setminus\Pi}$ in the front while maintaining the original order of the masked tokens for better predicting their positions. Moreover, the positional information of all masked tokens will be conditioned for better modeling of the full input sequence (e.g. the counts and positions of masked tokens in the sequence). 
The proposed BAT model is then adopted to predict the masked tokens as illustrated in Fig. \ref{fig_att}.

As shown in Fig. \ref{fig_att}, there is a masked sequence $X = \{x_1,[M],[M]$ $,x_4,[M]\}$ with length $L$ = 5 and positions $2,3,5$ being masked. After permutation and inserting the full mask tokens, we have the non-predicted tokens $(X_{\setminus M},M) = (x_1,x_4,[M],[M],[M])$ which provides the bidirectional context. For the predicted part, we have the input tokens $([M],x_2, x_3)$ to predict their corresponding next tokens \ie $(x_2,x_3,x_5)$. Here we use the mask token instead of $x_1$ to predict $x_2$ to encourage the leverage of positional information. 
We apply bidirectional modeling~\cite{devlin2018bert} to non-predicted tokens and autoregressive modeling to the predicted tokens to avoid future information leakage.
For example, while predicting $x_3$, the model could attend to $x_4$ in non-predicted tokens and meanwhile the previously `predicted' token $x_2$. Hence, we could capture bidirectional context and establish output dependency simultaneously with the proposed BAT.

\subsubsection{Transformer Architecture}
In this work, we adapt GPT~\cite{radford2019language} as our network architecture. The network is a decoder-only transformer that consists of $\mathcal{N}$  stacked decoder blocks. Given an intermediate embedding $H^n$ at the $n$-th layer, the decoder block can be formulated by:
\begin{align}
H^n & = H^n + \text{MA}(LN(H^n)) \\
H^{n+1} & = H^n + \text{MLP}(LN(H^n)),
\end{align}
where $MA$, $LN$ and $MLP$ stand for multihead self-attention, layer normalization, and fully-connected layers, respectively. For self-attention, we apply a customized mask to the $L\times L$ matrix of attention logits as illustrated in Fig.~\ref{fig_att}. At the final layer of the transformer, a learnable linear projection is employed to map $H^{\mathcal{N}}$ to logits, which parameterizes the conditional distribution for each pixel. 

During inference, we follow the raster-scan order to predict each masked token bidirectionally and autoregressively. We adopt a top-$\mathcal{K}$ sampling strategy to randomly sample from the $\mathcal{K}$ most likely next words. The predicted token is then concatenated with the input sequence as conditions for the generation of next masked token. This process repeats iteratively until all the masked tokens are sampled. Finally, the generated discrete sequence can be converted back to the RGB values with the aforementioned color palette.

\subsection{Texture Generator}

\subsubsection{Network Architecture}
As the inpainting diversity can be achieved by sampling the reconstructed structures $S$, we take the advantages of efficiency and texture representation capacity of CNNs to learn a deterministic mapping between low-resolution structures $S$ and high-resolution completed image $I_{out}$. The texture generator thus utilizes CNN layers and adversarial training to up-sample the reconstructed structures and replenish high-fidelity texture details by leveraging the styles of the valid pixels of input image $I_m$. In particular, we employ two encoders to encode the low-resolution structures and input images into two high-level CNN representations of the same dimension. We then concatenate them together as the input of a few consecutive residual blocks with different dilation rates. Finally, a SPADE~\cite{park2019semantic} generator is employed to incorporate the modulated style of input images and gradually up-sample the texture features to the target resolution. Meanwhile, all vanilla convolutions are replaced by gated convolution~\cite{yu2019free}.

\subsubsection{Loss Functions} The training of the texture generator is driven by the combination of several losses including a reconstruction loss, an adversarial loss, and a perceptual loss.
For clarity, we denote the texture generator as $G_t$, the ground truth as $I_{gt}$, and the completed image as $I_{out}$.
Firstly, a reconstruction loss $\mathcal{L}_{rec}$ between $I_{out}$ and $I_{gt}$ can be measured as follows:
\begin{align*}
    \mathcal{L}_{rec} = ||I_{out} - I_{gt}||_{1},
\end{align*}
Besides, a CNN-based discriminator $D$ together with an adversarial loss is employed to synthesize fine texture details. Specifically, the texture generator $G_t$ and discriminator $D$ are jointly trained with hinge loss \cite{isola2017image}, where the adversarial losses for the discriminator and generator are defined by:
\begin{align*}
    & \mathcal{L}_{adv}^{D} =  \mathbb{E}_{I_{gt}}[\textit{ReLU}(1-D(I_{gt})] 
                    + \mathbb{E}_{I_{out}}[\textit{ReLU}(1+D(I_{out})] \\
    & \mathcal{L}_{adv}^{G_t} =  - \mathbb{E}_{I_{out}}[D(I_{out})], \\
\end{align*}
Next, we penalize the perceptual and semantic discrepancy via the perceptual loss \cite{johnson2016perceptual} with a pretrained VGG-19 network:
\begin{align*}
    \mathcal{L}_{perc} = \sum_i{\lambda_i||\Phi_i(I_{out}) - \Phi_i(I_{gt})||_{1}} \\
    +\lambda_l||\Phi_l(I_{out}) - \Phi_l(I_{gt})||_{2},
\end{align*}
where $\lambda_i$ are balancing weights, $\Phi_i$ is the activation of $i$-th layer of the VGG-19 model (including \textit{relu1\_2}, \textit{relu2\_2}, \textit{relu3\_2}, \textit{relu4\_2} and \textit{relu5\_2}),
$\Phi_l$ represents the activation maps of \textit{relu4\_2} layer which mainly extracts semantic feature.
The texture generator is trained by optimizing the combination of aforementioned losses:
\begin{align*}
    \mathcal{L}_{G_t} = \underset{G_t}{min}\underset{D}{max}(\lambda_{rec}\mathcal{L}_{rec}+\lambda_{adv}\mathcal{L}_{adv}^{G_t}+\lambda_{perc}\mathcal{L}_{perc}),
\end{align*}
where $\lambda_{rec}$, $\lambda_{adv}$, and $\lambda_{perc}$ are empirically set at 1.0, 1.0 and 0.2, respectively, in our implementation.

\begin{figure*}[t]
\centering
\includegraphics[width=1.0\linewidth]{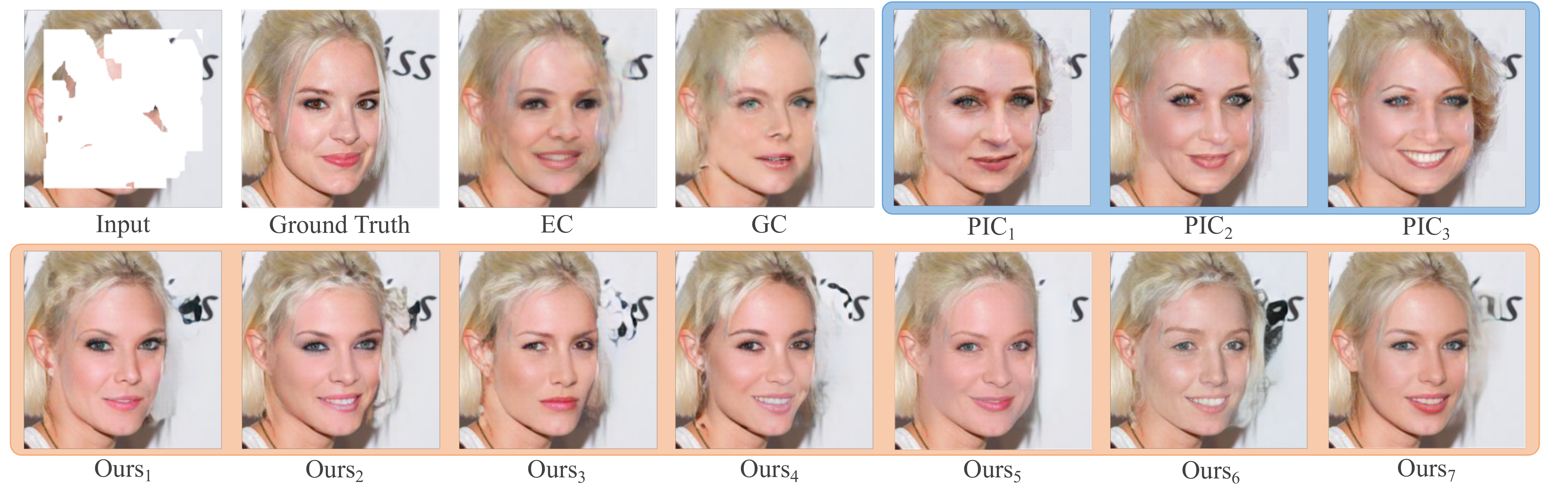}
\caption{
Qualitative comparisons of the proposed BAT-Fill with the state-of-the-art: BAT-Fill generates more realistic and diverse image inpainting over the dataset CelebA-HQ\cite{karras2017progressive} with irregular masks.
}
\label{fig_face}
\end{figure*}   

\begin{figure*}[t]
\centering
\includegraphics[width=1.0\linewidth]{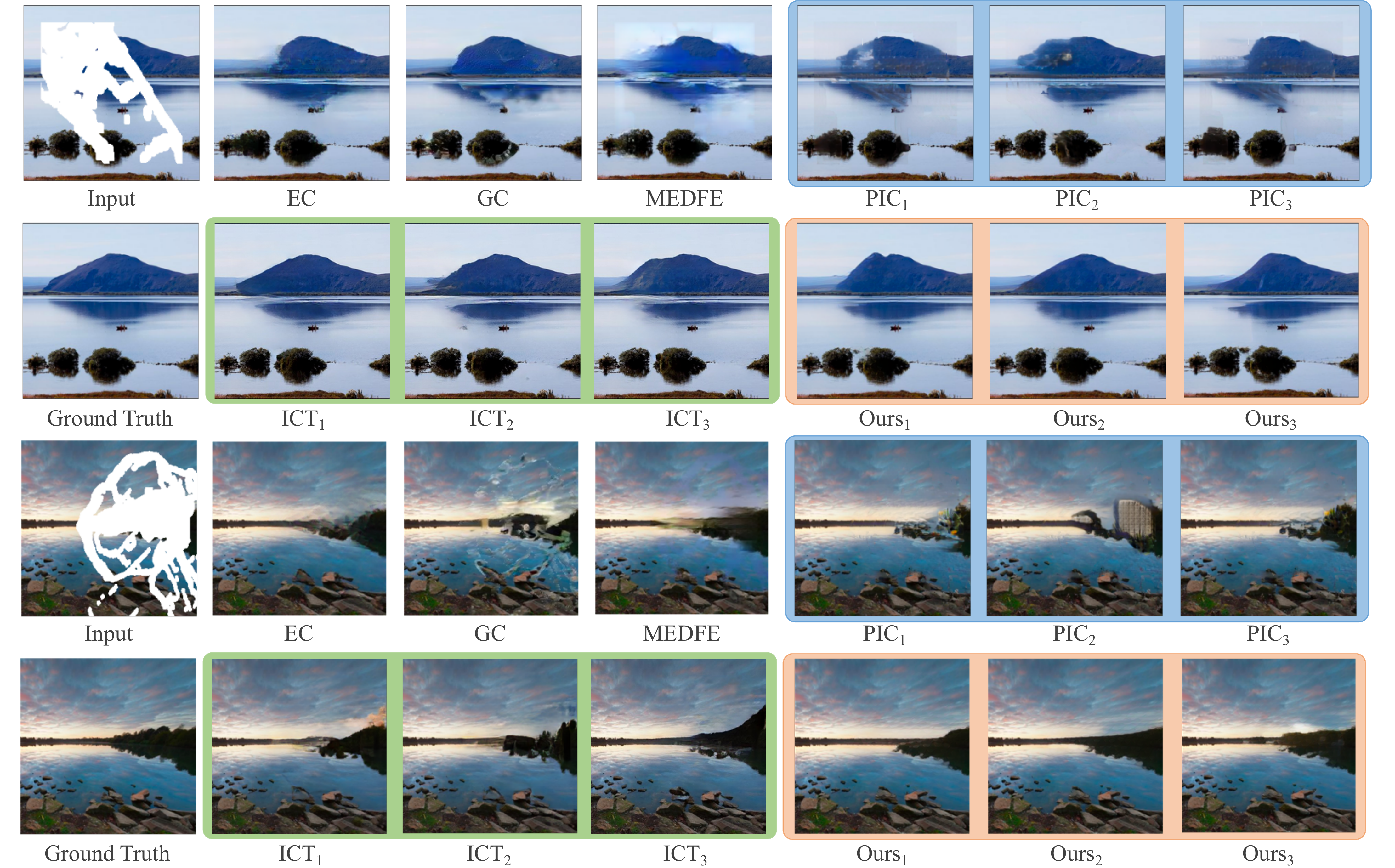}
\caption{
Qualitative comparison the proposed BAT-Fill with the state-of-the-art: BAT-Fill generates more realistic and diverse image inpainting over Places2~\cite{zhou2017places} with irregular masks.
}
\label{fig_view1}
\end{figure*} 

\begin{figure*}[t]
\centering
\includegraphics[width=1.0\linewidth]{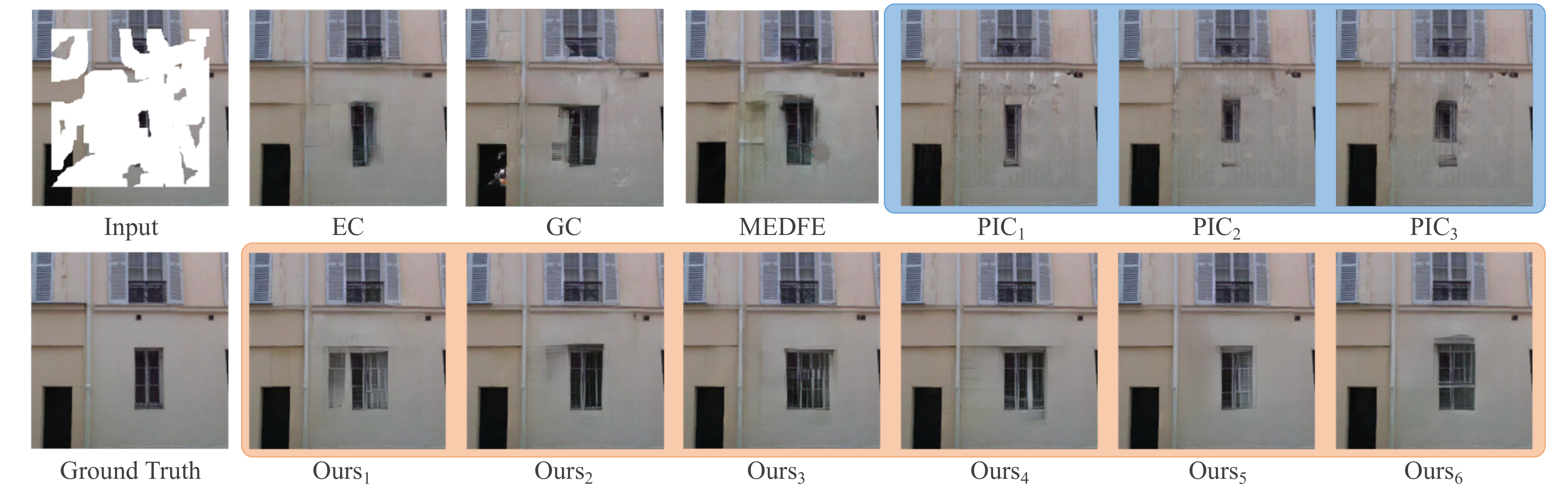}
\caption{
Qualitative comparison the proposed BAT-Fill with the state-of-the-art: BAT-Fill generates more realistic and diverse image inpainting over Paris StreetView~\cite{pathak2016context} with irregular masks.
}
\label{fig_view2}
\end{figure*} 

\section{Experiments}

\subsection{Experimental Settings}

\subsubsection{Datasets}
We conduct experiments over three public datasets that have different characteristics as listed:
\begin{compactitem}
    \item[--] CelebA-HQ \cite{karras2017progressive}: It is a high-quality version of the human face dataset CelebA \cite{liu2015deep} with 30,000 aligned face images. We follow the split in \cite{yu2019free} that produces 28,000 training images and 2,000 validation images, where 1,000 validation images are randomly sampled in evaluations.

    \item[--] Places2 \cite{zhou2017places}: It consists of more than 1.8M natural images of 365 different scenes. 
    We adopt the same 800 images from validation set with \cite{wan2021high} in evaluations.

    \item[--] Paris StreetView \cite{pathak2016context}: It is a collection of street view images in Paris, which contains 14,900 training images and 100 validation images.
\end{compactitem}

\subsubsection{Compared Methods}
We compare our method with a number of state-of-the-art methods as listed: 
\begin{compactitem}
    \item[--] GC \cite{yu2019free}: It is also known as DeepFill v2, a two-stage method that leverages gated convolutions.
    
    \item[--] EC \cite{nazeri2019edgeconnect}: It is a two-stage method that first predicts salient edges to guide the generation.  
    \item[--] MEDFE \cite{Liu2019MEDFE}: It is a mutual encoder-decoder that treats features from deep and shallow layers as structures and textures of an input image.
    
    \item[--] PIC \cite{zheng2019pluralistic}: It is a  probabilistically principled framework that leverages VAE to generate diverse image inpainting.
    
    \item[--] ICT \cite{wan2021high}: It is a diverse inpainting framework that combine the merits of transformers and CNNs for high-fidelity image inpainting. 
\end{compactitem}

\subsubsection{Evaluation Metrics}
We perform evaluations by using five widely adopted evaluation metrics: 1) Fr{\'e}chet Inception Score (FID) \cite{heusel2017gans} that evaluates the perceptual quality by measuring the distribution distance between the synthesized images and real images; 2) mean $\ell_1$ error; 3) peak signal-to-noise ratio (PSNR); 4) structural similarity index (SSIM) \cite{wang2004image} with a window size of 51; 5) Learned Perceptual Image Patch Similarity (LPIPS) \cite{zhang2018unreasonable} that evaluates the diversity of generated images. The average scores of LPIPS are calculated between random pairs of sampled inpainting results.

\subsubsection{Implementation Details}
The proposed method is implemented in PyTorch. The network is trained using $256 \times 256$ images with random irregular masks \cite{liu2018image}. The diverse-structure generator and texture generator are trained using $256 \times 256$ images with random irregular masks \cite{liu2018image}. We train the diverse-structure generator with AdamW~\cite{adamw} with $\beta_{1}=0.9$, $\beta_{2}=0.95$ and learning rate of 3e-4 following~\cite{chen2020generative}. For the texture generator, we use Adam optimizer \cite{kingma2014adam} with $\beta_{1}=0$ and $\beta_{2}=0.9$, and set the learning rate at 1e-4 and 4e-4 for the generator and discriminators, respectively. Learning rate decay is applied for the training of both networks, and the experiments are conducted on 4 NVIDIA(R) Tesla(R) V100 GPU.

\subsection{Quantitative Evaluation}
Extensive quantitative evaluations have been conducted over the three datasets with irregular masks~\cite{liu2018image}. The irregular masks in the experiments are categorized according to the mask ratios, and an additional category `random' is evaluated which randomly samples masks with ratios varying from 20\% to 60\%. The performance of the compared methods was acquired by using the publicly available pre-trained models or implementation codes.
\footnote{\url{https://github.com/JiahuiYu/generative_inpainting}} \footnote{\url{https://github.com/knazeri/edge-connect}} \footnote{\url{https://github.com/KumapowerLIU/Rethinking-Inpainting-MEDFE}}
\footnote{\url{https://github.com/lyndonzheng/Pluralistic-Inpainting}}. 

We compare the proposed method with both deterministic and diverse image inpainting methods. Note that all reference metrics such as $\ell_1$, SSIM ,and PSNR are in favor of deterministic inpainting methods where the prediction is directly compared with the ground truth. 
Different from PIC~\cite{zheng2019pluralistic} that unitizes its discriminator to sort the results, our method adapts the top-50 sampling strategy and use all random samples for fair comparisons, which means our method directly generates the stochastic inpainting without the additional filtering.

Table~\ref{tab_psv} shows the inpainting performance over the dataset Paris StreetView~\cite{pathak2016context}. Compared with deterministic methods GC, EC, and MEDFE, the proposed method achieves the best FID scores over different mask ratios and consistently outperforms the diverse inpainting method PIC in both inpainting quality (FID) and inpainting diversity (LPIPS). In addition, Table~\ref{tab_quant2} shows the inpainting performance over CelebA-HQ~\cite{karras2017progressive} and Places2~\cite{zhou2017places}. For CelebA-HQ, our method consistently outperforms all compared methods, especially in FID scores. For Places2, our method achieves comparable performance with deterministic methods in all evaluation metrics, and it generally outperforms them in FID scores. In addition, the numerical results of BAT-Fill suggest a clear superiority over the diverse inpainting method PIC~\cite{zheng2019pluralistic}, and better FID scores than ICT~\cite{wan2021high}.

\renewcommand\arraystretch{1.2}
\begin{table*}[t]
\caption{
Quantitative comparison of the proposed BAT-Fill with state-of-the-art methods over CelebA-HQ \cite{karras2017progressive} and Places2 \cite{zhou2017places} validation images (1,000) with irregular masks \cite{liu2018image} ($\ast$ denotes that we trained the model based on official implementations, $\dagger$ denotes the results are copied from \cite{wan2021high}). For each metric, the best score is highlighted in \textbf{bold}, and the best score for diverse inpainting methods (\ie PIC~\cite{zheng2019pluralistic} and Ours) is highlighted in \underline{underline}.
}
\label{tab_quant2}
\small 
\renewcommand\tabcolsep{2pt}
\centering 
\scalebox{0.92}{
\begin{tabular}{l|c||ccc||ccc||ccc||ccc} \hline
\multirow{2}{*}{\textbf{Methods}} & \multirow{2}{*}{\textbf{Dataset}} &
\multicolumn{3}{c||}{\textbf{FID}$\downarrow$} & 
\multicolumn{3}{c||}{$\boldsymbol{\ell_1(\%)}\downarrow$} &
\multicolumn{3}{c||}{\textbf{PSNR}$\uparrow$} &
\multicolumn{3}{c}{\textbf{SSIM}$\uparrow$}
\\
\cline{3-14}
& & 20-40\% & 40-60\% & Random
& 20-40\% & 40-60\% & Random
& 20-40\% & 40-60\% & Random
& 20-40\% & 40-60\% & Random
\\\hline
\textbf{EC$\ast$ \cite{nazeri2019edgeconnect}} & \multirow{4}{*}{CelebA-HQ~\cite{karras2017progressive}} & 9.06 & 16.45 & 12.46 & 2.19 & 4.71 & 3.40 & 26.60 & 22.14 & 24.45 & 0.923 & 0.823 & 0.877  \\
\textbf{GC \cite{yu2019free}} & &14.12 & 22.80 & 18.10 & 2.70 & 5.19 & 3.88 & 25.17 & 21.21 & 23.32 & 0.907 & 0.805 & 0.858  \\
\textbf{PIC \cite{zheng2019pluralistic}} & &10.21 & 18.92 & 14.12 & 2.50 & 5.65 & 4.00 & 25.92& 20.82 & 23.46 & 0.919 & 0.780 & 0.852  \\

\textbf{Ours} & & \textbf{\underline{6.32}} & \textbf{\underline{12.50}}&  \textbf{\underline{9.33}}& \textbf{\underline{1.91}} & \textbf{\underline{4.57}}
& \textbf{\underline{3.18}} &\textbf{\underline{27.82}} & \textbf{\underline{22.40}}&  \textbf{\underline{25.21}}& \textbf{\underline{0.944}} & \textbf{\underline{0.834}}& \textbf{\underline{0.890}}
  \\\hline
 \textbf{EC$\dagger$ \cite{nazeri2019edgeconnect}} & \multirow{6}{*}{Places2~\cite{zhou2017places}} & 25.64 & 39.27 & 30.13 & 2.20 & 4.38 & 2.93 & 26.52 & 22.23 & 25.51 & 0.880 & \textbf{0.731} & 0.831  \\
\textbf{GC$\dagger$ \cite{yu2019free}} & &24.76 & 39.02 & 29.98 & \textbf{2.15} & 4.40 & 2.80 & \textbf{26.53} & 21.19 & 25.69 & \textbf{0.881} & 0.729 & \textbf{0.834}   \\
\textbf{MEDFE$\dagger$ \cite{Liu2019MEDFE}} & &26.98 & 45.46 & 31.40 & 2.24 & 4.57 & 2.91 & 26.47 & \textbf{22.27} & 25.63 & 0.877 & 0.717 & 0.827  \\
\textbf{PIC$\dagger$ \cite{zheng2019pluralistic}} & &26.39 & 49.09 & 33.47 & 2.36 & 5.07 & 3.15 & 26.10 & 21.50 & 25.04 & 0.865 & 0.680 & 0.806  \\
\textbf{ICT$\dagger$ \cite{wan2021high}} & &21.60 & 33.85 & 25.42 & 2.44 & \textbf{\underline{4.31}} & \textbf{\underline{2.67}} & \underline{26.50} & \underline{22.22} & \textbf{\underline{25.79}} & \underline{0.880} & \underline{0.724} & \underline{0.832}  \\
\textbf{Ours} & & \textbf{\underline{17.78}} & \textbf{\underline{32.55}} &  \textbf{\underline{22.16}}& \textbf{\underline{2.15}} & 4.64 & 2.84 & 26.47 & 21.74 & 25.69 & 0.879 & 0.704 & 0.826 

  \\\hline
\end{tabular}}
\end{table*}

\subsection{Qualitative Evaluations}

Figs. \ref{fig_face}, \ref{fig_view1}, and \ref{fig_view2} show the qualitative comparisons between BAT-Fill and the state-of-the-art image inpainting methods over the validation set of CelebA-HQ \cite{karras2017progressive}, Places2 \cite{zhou2017places} and Paris StreetView \cite{pathak2016context}, respectively.

We first evaluate and compare BAT-Fill with EC \cite{nazeri2019edgeconnect}, GC \cite{yu2019free}, and PIC \cite{zheng2019pluralistic} on CelebA-HQ \cite{karras2017progressive} which contains facial images with similar semantics. 
As shown in Fig. \ref{fig_face}, though EC \cite{nazeri2019edgeconnect} and GC \cite{yu2019free} can synthesize complete facial images with reasonable semantics, they tend to generate distorted facial structures and artifacts in the missing regions which degrades 
inpainting greatly. In addition, EC \cite{nazeri2019edgeconnect} and GC \cite{yu2019free} can only generate deterministic inpainting, which limits their applicability clearly.
Both PIC \cite{zheng2019pluralistic} and BAT-Fill can generate diverse inpainting. However, the PIC generated images share similar makeups and facial features and thus have limited diversity.
As a comparison, the BAT-Fill generated facial images vary across a wide range of makeups and facial features and contain much less artifacts, 
demonstrating that BAT-Fill can produce more diverse and realistic inpainting.

Next, we evaluate and compare BAT-Fill with EC \cite{nazeri2019edgeconnect}, GC \cite{yu2019free}, MEDFE \cite{Liu2019MEDFE}, and PIC \cite{zheng2019pluralistic} on the datasets Places2 \cite{zhou2017places} and Paris StreetView \cite{pathak2016context} where images have various semantics. In addition, visual comparison with ICT \cite{wan2021high} is conducted over Places2 \cite{zhou2017places} dataset.
As shown in Fig. \ref{fig_view1}, EC \cite{nazeri2019edgeconnect}, GC \cite{yu2019free} and MEDFE \cite{Liu2019MEDFE} tend to generate blurs and even corrupted texture in the inpainting images. The PIC \cite{zheng2019pluralistic} synthesized images suffer from unreasonable semantics, obvious artifacts, and limited diversity. 
Both ICT \cite{wan2021high} and BAT-Fill achieved realistic image inpainting with much less artifacts and better diversity compared with other methods.
For Paris StreetView \cite{pathak2016context}, BAT-Fill produced more diverse and plausible results than the PIC \cite{zheng2019pluralistic}, and meanwhile achieved comparable or even better inpainting quality compared with the deterministic methods.

\renewcommand\arraystretch{1.2}
\begin{table}
\caption{
Quantitative comparison of the proposed BAT-Fill with state-of-the-art methods over Paris StreetView \cite{pathak2016context} validation images (100) with irregular masks \cite{liu2018image} ($\ast$ denotes that we trained the model based on official implementations). For each metric, the best score is highlighted in \textbf{bold}, and the best score for diverse inpainting methods (\ie PIC~\cite{zheng2019pluralistic} and Ours) is highlighted in \underline{underline}.
}
\label{tab_psv}
\renewcommand\tabcolsep{2pt}
\centering 
\small
\scalebox{0.9}{
\begin{tabular}{l|c||ccccc} \hline
\multirow{2}{*}{\textbf{Metrics}} & \multirow{2}{*}{\shortstack{\textbf{Mask}\\ \textbf{Ratio}}} &
\multicolumn{3}{c}{\textbf{Methods}}
\\
\cline{3-7} &  &EC~\cite{nazeri2019edgeconnect} &GC$\ast$~\cite{yu2019free} & MEDFE~\cite{Liu2019MEDFE} & PIC~\cite{zheng2019pluralistic}
& Ours
\\\hline
{\textbf{FID}$\downarrow$} & \multirow{5}{*}{20-40\%}  & 42.81&71.02& 36.84 & 56.83 & \textbf{\underline{36.19}}   \\
{$\boldsymbol{\ell_1(\%)}\downarrow$} &  & 2.63& 3.56 & \textbf{2.29} & 3.43 & \underline{2.70}  \\
{\textbf{PSNR}$\uparrow$} & &26.76  & 23.95& \textbf{27.64} & 24.80 &  \underline{26.52}  \\
{\textbf{SSIM}$\uparrow$} & & 0.874 & 0.796& \textbf{0.898} & 0.817 & \underline{0.864} \\
{\textbf{LPIPS}$\uparrow$} & &N/A &N/A &N/A & 0.046 &  \textbf{\underline{0.076}}  \\
\hline
{\textbf{FID}$\downarrow$} & \multirow{5}{*}{40-60\%}  & 72.78 &98.32 & 77.26 & 90.91 &  \textbf{\underline{64.20}}  \\
{$\boldsymbol{\ell_1(\%)}\downarrow$} & &\textbf{5.18} & 6.31& 5.54 & 7.47 &  \underline{5.83}  \\
{\textbf{PSNR}$\uparrow$} & &\textbf{22.77}&20.83 & 22.01 & 20.12 & \underline{21.89}  \\
{\textbf{SSIM}$\uparrow$} & &\textbf{0.712}&0.631 & 0.704 & 0.570 &  \underline{0.678}  \\
{\textbf{LPIPS}$\uparrow$} & &N/A&N/A  & N/A & 0.127 & \textbf{\underline{0.147}} \\
\hline
{\textbf{FID}$\downarrow$} & \multirow{5}{*}{Random}  & 55.29& 84.16 & 54.99 & 72.16 & \textbf{\underline{48.19}}   \\
{$\boldsymbol{\ell_1(\%)}\downarrow$} & &3.63&4.64 & \textbf{3.58} & 4.94 &  \underline{3.96}  \\
{\textbf{PSNR}$\uparrow$} & &25.04 &22.61 & \textbf{25.24} & 22.97 &  \underline{24.50} \\
{\textbf{SSIM}$\uparrow$} & &0.806 &0.727 & \textbf{0.818} & 0.718 & \underline{0.786} \\
{\textbf{LPIPS}$\uparrow$} & &N/A&N/A & N/A & 0.082 & \textbf{\underline{0.106}} \\
\hline
\end{tabular}}
\end{table}

\subsection{Ablation Study}

We study the effectiveness of the proposed BAT by conducting ablation studies over Paris StreetView~\cite{park2019semantic}. In the ablation study, we remove the two key components from BAT respectively, which result in two models: 1) w/o bidirectional context, where we will get the same objective with the autoregressive model that predicts the missing tokens by conditioning on previous tokens with unidirectional attention; 2) w/o autoregressive model, where the model is equivalent to MLM that independently reconstruct the missing tokens. To measure the diversity of MLM, we employ a Gibbs sampling to iteratively sample tokens and place the predicted tokens into the original sequence instead of directly output all the predicted tokens. 
For a fair comparison, we apply the same irregular masks (mask ratios 40-60\%) on the same low-resolution images ($32\times32$) from the validation set of Paris StreetView~\cite{park2019semantic}. After predicting the same inputs, the reconstructed structures of each model are evaluated without applying the texture generator.

As shown in Table~\ref{tab_as1}, using AR greatly degrades the quality of the reconstructed structures, and the high diversity measured by LPIPS is also largely attributed to the poor reconstruction quality. MLM performs reasonably well as it exploits the bidirectional context for inpainting. However, the proposed BAT clearly outperforms in reconstruction quality which is mainly reflected by FID, and it achieves comparable diversity as reflected by LPIPS. This is mainly because BAT models the output dependency to align the future predictions with previously predicted tokens and improves the consistency of the reconstructed structures. Overall, the ablation study demonstrates that the proposed BAT addresses the constraints of the AR and MLM effectively.

\renewcommand\arraystretch{1.2}
\begin{table}
\caption{
Ablation study of the proposed BAT over Paris StreetView \cite{park2019semantic} validation set (100) with irregular masks \cite{liu2018image} and mask ratios of 40\%-60\%.}
\label{tab_as1}
\small
\begin{center}
\scalebox{0.85}{
\begin{tabular}{|l|ccccc|}
\hline
Models & FID$\downarrow$ & $\ell_1(\%)\downarrow$ & PSNR$\uparrow$ & SSIM$\uparrow$ & LPIPS$\uparrow$\\
\hline\hline
w/o bidirectional &59.60&8.599&19.32&0.518 & \textbf{0.0518} \\

w/o autoregressive &49.62&6.38&22.01&0.655 & 0.0304\\

\hline
Ours  &\textbf{40.91}&\textbf{5.76}&\textbf{23.01}&\textbf{0.714} & 0.0301\\


\hline
\end{tabular}}
\end{center}

\end{table}  

\section{Conclusion}

This paper presents BAT-Fill, a novel image inpainting framework that achieves realistic and diverse inpainting by leveraging the autoregressive transformers with their powerful long-dependency modeling capacity. To improve the quality and diversity of inpainting, we propose a novel bidirectional and autoregressive transformer (BAT) to model the bidirectional context and output dependency simultaneously. Extensive experiments show that BAT-Fill achieves superior image inpainting in terms of both quality and diversity. 
Moving forward, we will explore the feasibility of adapting our idea to other image recovery or generation tasks by replacing the non-predicted part of BAT with other conditions such as semantic label, edge, and pose.

{\small
\bibliographystyle{ieee_fullname}
\bibliography{reference}
}

\end{document}